\documentclass{ecai}
\usepackage{times}
\usepackage{graphicx}
\usepackage{latexsym}

\begin{document}

\title{Dual Local-Global Contextual Pathways for Recognition in Aerial Imagery}

\author{Alina Marcu \and Marius Leordeanu
\institute{Autonomous Systems and University Politehnica of Bucharest, Romania,
emails: alymarcu91@gmail.com \and leordeanu@gmail.com} }

\maketitle
\bibliographystyle{ecai}

\begin{abstract}
  \textit{Visual context is important in object recognition and it is still an open problem in computer vision.
  Along with the advent of deep convolutional neural networks (CNN), using contextual information with such systems starts to receive attention in the literature.
  At the same time, aerial imagery is gaining momentum.
  While advances in deep learning make good progress in aerial image analysis, this problem still poses many great challenges.
  Aerial images are often taken under poor lighting conditions and contain low resolution objects, many times occluded by trees or taller buildings.
  In this domain, in particular, visual context could be of great help, but there are still very few papers that consider context in aerial image understanding.
  Here we introduce context as a complementary way of recognizing objects. We propose a dual-stream deep neural network model that
  processes information along two independent pathways, one for local and another for global visual reasoning. The two
  are later combined in the final layers of processing. Our model learns to combine local object appearance as well as information from the larger scene at the same time and in a complementary way, such that together they form a powerful classifier. We test our dual-stream network on the task of segmentation of buildings and roads in aerial images and obtain state-of-the-art results on the Massachusetts Buildings Dataset. We also introduce two new datasets, for buildings and road segmentation, respectively, and study the relative importance of local appearance vs. the larger scene, as well as their performance in combination. While our local-global model could also be useful in general recognition tasks, we clearly demonstrate the effectiveness of visual context in conjunction with deep nets for aerial image understanding.}
\end{abstract}

\section{INTRODUCTION}

Object recognition in aerial imagery is enjoying a growing interest today,
due to the recent advancements in computer vision and deep learning,
along with important improvements in low-cost high performance GPUs.
The possibility of accurately recognizing different types of objects in aerial images,
such as buildings, roads, vegetation and other categories,
could greatly help in many applications, such as
creating and keeping up-to-date maps,
improving urban planning, environment monitoring and disaster relief.
Besides the practical need for accurate aerial image interpretation systems, this domain also offers specific
scientific challenges to the computer vision domain. Aerial images require the recognition
of very small objects, seen from above under difficult lighting conditions, which are sometimes occluded or only partially seen.
One point we make in our paper is that visual context is vital for accurate recognition in such cases.

We study the importance of visual context and propose a dual-stream deep convolutional neural network
that combines local appearance with more global scene information in a complementary way.
Thus, the object is seen both as a separate entity
from the perspective of its own appearance, but
also as a part of a larger scene,
which acts as its complement and implicitly contains information about it.
Our local-global model offers a dual view of the object, with one processing pathway, based on a properly
adjusted VGG-Net~\cite{simonyan2014very} that focuses on local, object level information, and a second one,
using a modified AlexNet~\cite{krizhevsky2012imagenet},
which considers information from the larger area around the object of interest. The two pathways are then joined into
a final subnet composed of three fully-connected (FC) layers, where the intermediate results are combined and potential disagreements
resolved for a final output. We formulate the problem as one of segmentation in the sense of finding an accurate shape for the object of interest.
Our combined network, which we term LG-Seg, is trained jointly, end-to-end.

We bring two main contributions. First, we demonstrate experimentally that larger visual
context is important for semantic segmentation in aerial images
and show superior performance to current state-of-the-art methods on the Massachusetts Buildings Dataset.
Demonstrating the importance of the larger visual scene context in aerial imagery is relevant, since
current techniques in aerial imagery focus only on local object appearance.
Second, we propose a novel dual-stream deep CNN architecture, with two processing pathways, one for local and the other for global image interpretation.
We show in our experiments that the two pathways learn to process information complementarily in order
to obtain an improved output.

\section{VISUAL CONTEXT AND AERIAL IMAGERY}

Context could play a fundamental role in aerial image understanding, especially in cases of poor resolution, poor lighting or occlusion.
For example, a square in the middle of a residential area could be more confidently labeled as a building than in the middle of a road or a large body of water. Thus, the same square, with exactly the same appearance, could be \emph{seen} differently.

There is a lot of relevant work for various computer vision problems that study and use contextual information.
Earlier approaches used global
scene features for object recognition~\cite{torralba2003contextual, oliva2007role, torralba2006contextual}.
Other works used only the immediate
neighborhood of an object, which often provides strong cues for image recognition or tracking in video
~\cite{zhu2015segdeepm, collins2005online, leordeanu2016labeling}.
There are many different techniques and tasks related to the use of context in vision, such as methods based on CRFs~\cite{yao2012describing} or
algorithms for inferring the 3D layout of objects and orientations of surfaces~\cite{hoiem2008putting}. Other ideas use contextual relationships
between objects, such as co-occurrences between different categories~\cite{rabinovich2007objects}. The presence of different object detectors in the vicinity
of the box of interest is also known to increase recognition performance~\cite{felzenszwalb2010object}. Other methods based on relationships between
objects include modeling spatial relations as a structured prediction problem~\cite{desai2011discriminative}.
One successful approach in semantic segmentation, known as autocontext~\cite{tu2010auto},
uses classifier outputs from one level of image interpretation as contextual
inputs to a higher level of abstraction.
Context could be understood in many forms, going from reasoning about objects against the global scene~\cite{torralba2010using}
to looking at more precise spatial and temporal relationships and interactions between different
object categories~\cite{leordeanu2014thoughts}. One relevant example is work~\cite{choi2010exploiting}
that combines both spatial relations to other objects as well as global scene context.

It is not yet known what is the best way to combine object relationships and global information for contextual reasoning.
Deep neural networks are an interesting choice for modeling context, as they
process information from one level of abstraction to the next. They use single, discrete neurons, which combined with different ways of pooling
could model "detections" of deferent features, object parts or even whole concepts, at different levels of abstraction. Thus they relate to methods
using object detectors for extra contextual cues.
By using many such neurons, with soft responses over potentially large fields, they could also model
global image statistics - connecting to literature using whole image contextual features.
By reasoning in a hierarchical manner they also offer the possibility of integrating information from one level as contextual input to the next,
relating to approaches using autocontext.
Therefore, deep nets seem to offer the right environment for designing effective
architectures for using and studying visual context. Their recent success in computer vision on various tasks
~\cite{krizhevsky2012imagenet,lecun1989backpropagation,simonyan2014very,girshick2015fast,szegedy2015going,zhang2015improving}
encouraged researchers to start testing different approaches for using context in conjunction with CNNs.
Such systems, combining context with deep networks,
were proposed for action classification~\cite{gkioxari2015contextual}, segmentation by modeling
CRFs \cite{zheng2015conditional} with recurrent networks and object detection by
training contextual networks over nearby bounding box regions~\cite{zhu2015segdeepm,gidaris2015object}.
Other recent work models person context in order to improve detection of objects that are used by or
related to people~\cite{gupta2015exploring}. Another recent architecture is designed for integrating local and holistic information
for human pose estimation~\cite{fan2015combining}.
Note that
research in using visual context for object detection is also limited by current image datasets, such as PASCAL VOC Dataset~\cite{everingham2010pascal},
in which objects occupy a large part of the image.
Different from~\cite{zhu2015segdeepm,gidaris2015object} our proposed deep architecture is based on a dual-stream network, each pathway having
its own different architecture, centered on the object but looking over different image areas: one considering local information and the other taking into account
a much larger region. As we show in our experiments, the two pathways learn by themselves to process the object and its surroundings in two complementary ways, one for finer shape segmentation and the other for reasoning about the larger context.

Different from previous work, we study context in the domain of aerial imagery, where objects are
relatively small and it is easy to include larger areas as input.
In aerial imagery most traditional approaches are based on multiple cues extracted from the image such as
color bands, gradients, histograms or certain geometric features.
Objects are first detected using each feature independently and then, by applying a decision fusion method~\cite{senaras2013building},
the results from previous features are combined. The method in~\cite{sirmaccek2008building} for detection of buildings in aerial imagery,
extracts several features from the main scene in order to highlight the areas of interest containing the buildings, then uses invariant color features, edge and shadow information in order to segment their exact shape. Other work selects the most discriminative features for semantic classification in aerial imagery using boosting
~\cite{tokarczyk2015features}. There is also recent work~\cite{mattyus2015enhancing} that combines satellite aerial images available online with ground truth labels from OpenStreeMap (OSM) for learning to enhance road maps. Authors use some weak context features based on differences in mean pixels intensities between the
road area and its background, within a Markov Random Fields formulation.
Very few approaches in aerial image analysis use CNNs, with improved results
~\cite{saito2015building,MnihThesis}.

Our main contribution over the prior work is to show that contextual information is important for accurate object recognition in aerial images and also provide a novel
dual-stream architecture, based on deep convolutional neural networks, which learns in parallel to recognize objects from two complementary views, one from the local level of object appearance and the other from the contextual level of the scene.

\section{PRELIMINARY WORK AND INTUITION}

\begin{figure*}[t!]
\centering
\includegraphics[width=17cm]{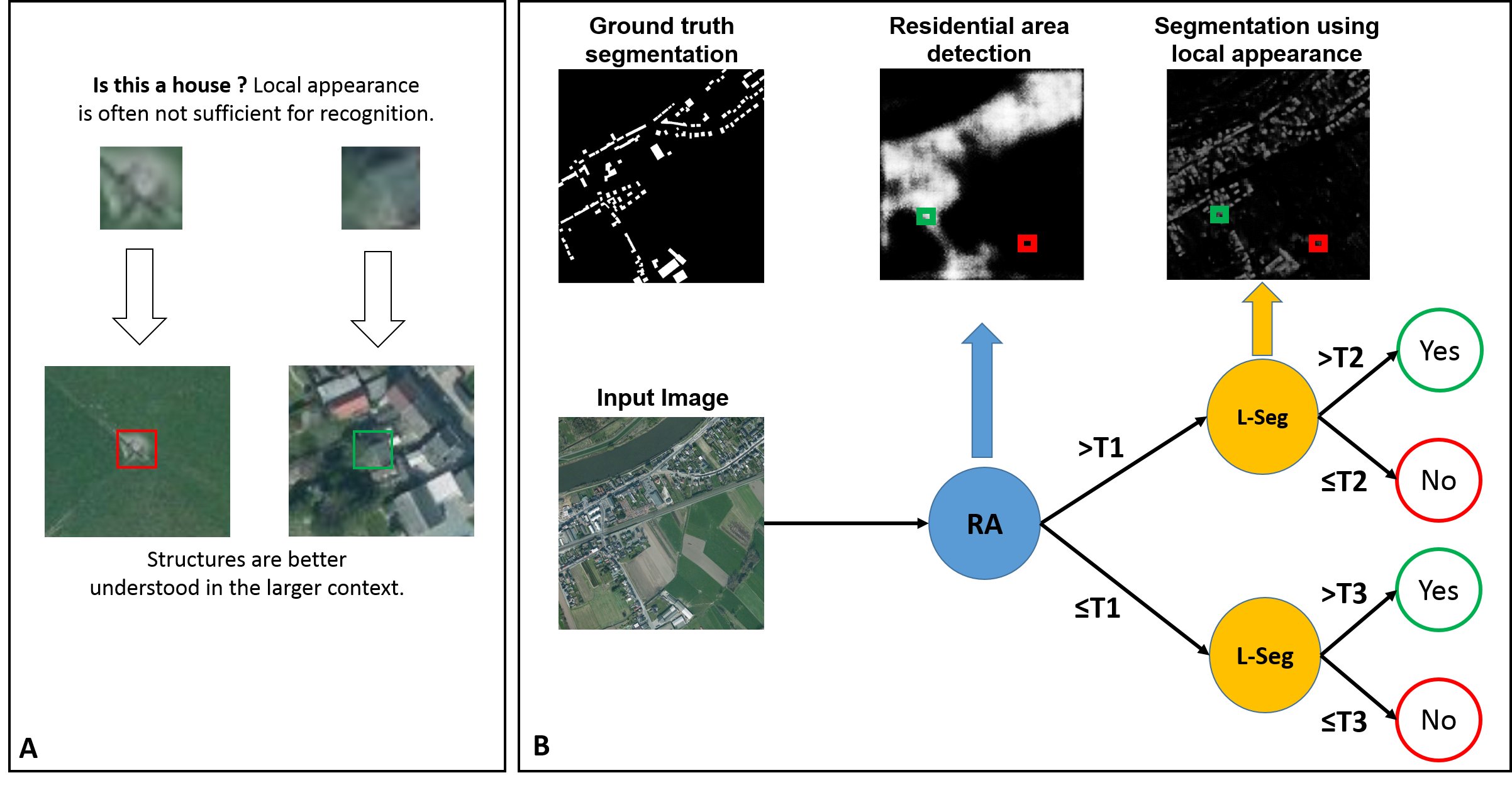}
\caption{\textbf{A}: Local appearance is often not sufficiently informative for segmentation in low-resolution aerial images.
The larger context could provide vital information even for highly localized tasks such as fine object segmentation: the exact shape of the house
in the example on the right is better perceived when looking at the larger residential area, which contains other houses of similar shapes and orientations.
Thus, local structure could be better interpreted in the context of the larger scene.
\textbf{B}: Our initial model for residential area detection (RA) has
poor localization but low false positive rate within a larger neighborhood. RA
can be effectively combined, in a simple classification tree, with the local semantic segmentation model (L-Seg),
which has higher localization accuracy but relatively high false positive rate.
Note how the output from RA can be used in order to filter out the houses hallucinated by the local L-Seg model.
Best viewed on the screen.
}
\label{fig:motivation}
\end{figure*}

Let us look at Figure~\ref{fig:motivation} A. We present two local patches and their larger scene context. By looking at the patches only,
it appears that local appearance is not sufficient for confidently recognizing the presence and the shape of a house.
In fact, from the local patch alone, the example on the left seems to be more likely to
belong to a house than the one on the right. When we consider the larger contextual neighborhood, the house roof
is more clearly perceived in
the second case, in which the larger residential area contributes in an important way to the local perception.

Geometric grouping cues such as agreements of houses' orientations and similar appearances in the larger residential area increase
the chance that we are indeed looking at a house and also help "seeing" its shape better.
In the case on the left, the contextual alignment of the diagonals in the larger region of grass lowers the
possibility that we are indeed looking at a house.
We argue
that larger contextual influences are not only important for determining the presence or absence of a certain object class,
but are also important for a more accurate perception of shape. Our experiments in Section \ref{sec:experiments}  also confirm this fact.

\subsection{Buildings vs. Residential Regions}
We consider the problem of finding the shapes of buildings in an aerial image. We treat the task from two perspectives,
considering both their local appearance as well as information from the larger scene containing them.
We are interested to study
the role of context on this task, as buildings have various shapes and appearances and are representative
for most aerial images.
We employ two models based on CNNs.
First, a local deep neural network, based on the state-of-the-art VGG-Net, is trained to output $16 \times 16$ patches of pixelwise labels,
with values between $0$ and $1$,
in order to predict the presence or absence of a building at a given pixel. At test time the image is
divided into a disjoint set of patches, on a grid, and each patch is classified independently. The end result becomes a segmentation of the entire image, with
white areas belonging to building pixels.
The input to the network is a larger 64 x 64 patch that, in the case of smaller houses, often
contains little surrounding background information. This network is thus trained to detect and segment houses
(output their exact shapes) using mostly local information. We will refer to it as the local L-Seg network.

In order to study the role of the larger context, we employ a wider (with larger filters and input)
but shallower architecture based on AlexNet, which takes as input a $256 \times 256$
image patch ($16 \times$ larger in area than the input to L-Seg)
centered at the same location. This second model is not trained for accurate shape prediction, but only to output a single binary variable - whether the input patch
belongs to a residential area or not.
In our case, a large 256 x 256 patch is
considered to be residential if it contains at least $15$ houses. This is
a moderate number for such patches in an image with 1 sq. m per pixel.
We tested with different numbers
in the $5-30$ range, with similar results.
For training, the non-residential patches were
not allowed to contain any buildings.
Our goal was to see whether the two models, trained completely separately on two different tasks, one for accurate shape
segmentation and the other for simple binary classification
could be later combined for improved performance.

\subsection{Initial experiments}
\begin{figure}[t!]
\centering
\includegraphics[height=4.8cm]{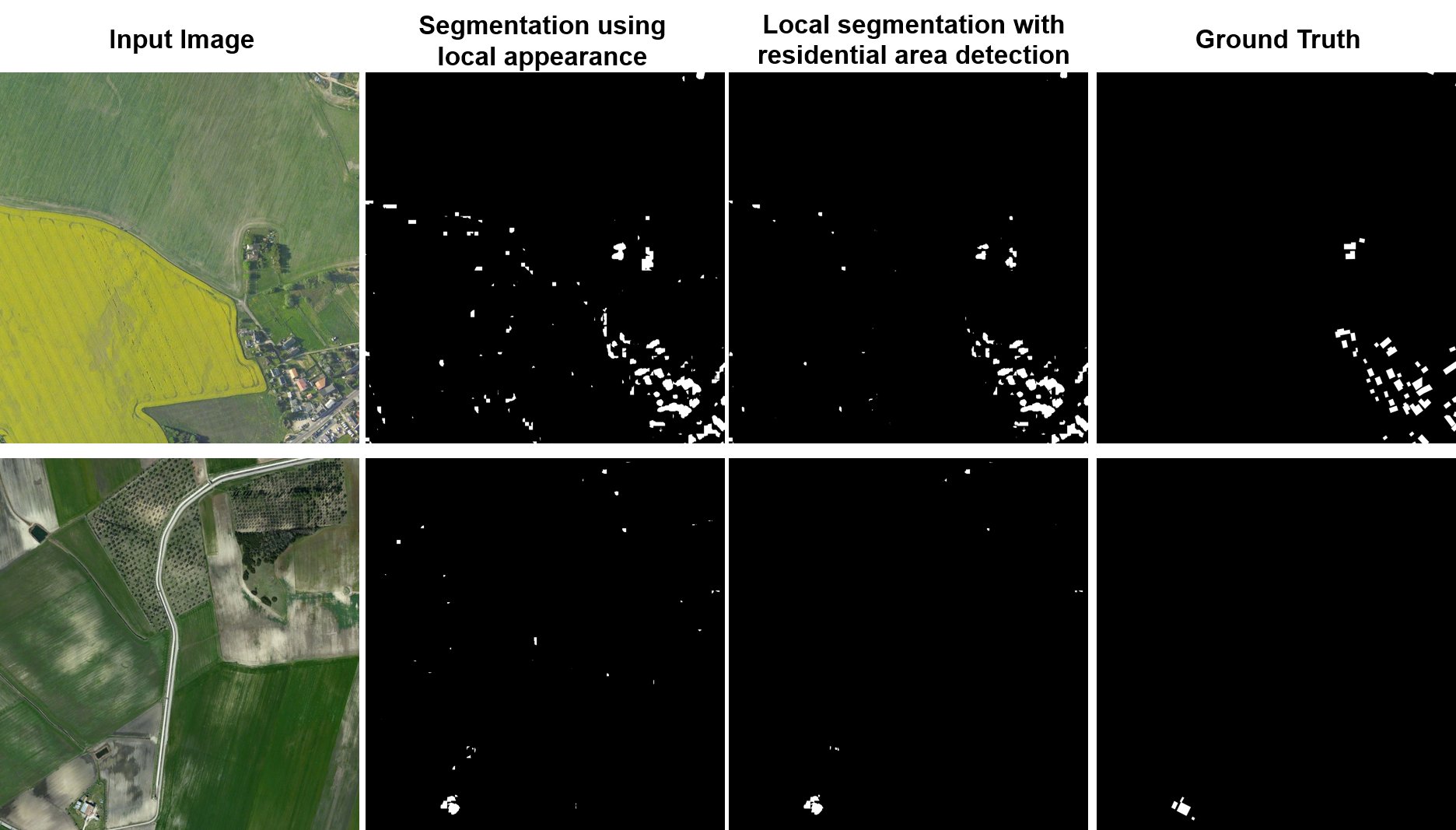}
\caption{Segmentation of buildings by the combined tree model (third column) vs. the local segmentation alone (second column),
without the residential area detector (RA). Note the decrease in false positives
in regions with very low residential density, showing the usefulness of the RA classifier.}
\label{fig:tree_classifier_results}
\end{figure}

We found that the two models can be effectively joined into a classifier tree, with the residential area classifier (RA) acting as a filter that
reduces the false positive rate of the local buildings shape segmenter L-Seg (Figure \ref{fig:tree_classifier_results}).
While the L-Seg CNN segments disjoint patches on a grid, the RA classifier gives single labels to those patches.
A dense pixelwise residential area classification could be obtained by interpolation.
The tree model is formed (Figure \ref{fig:motivation} B) by putting the RA classifier at a first node and the L-Seg model at the leaves.
Depending on how the first node classifies the patch, the leaves will classify it using different thresholds. Consequently,
if a patch is classified as residential by RA, the segmenter L-Seg will be more likely to detect buildings than otherwise.

The tree is controlled by the two models with three different thresholds T1, T2 and T3. T1 is applied to the RA classifier, while T2 and T3 control the precision of the L-Seg leaves. The three parameters are optimized in sequence, until convergence, as follows: before the first iteration, the thresholds are chosen independently to maximize the F-measure of the two classifiers. Then, each threshold is optimized in turn, while the other two are kept fixed. The F-measure is thus improved from $59.8 \%$ to $60.6\%$ on the European Buildings Dataset (presented in detail in Section \ref{sec:experiments}).
Note that these numbers are relatively low compared to the ones from the experimental section, because on these initial experiments we stopped the training of the CNNs
relatively early, before complete convergence. Also, for evaluation we did not use the relaxation of three pixels which we applied later, in order to compare with other methods. At this point all we are interested in,
is whether a residential area detector can be combined effectively and in a simple way with a local buildings segmenter.
We should also note that the overall quantitative improvement of $0.8\%$
is an average value over all pixels in the test set. It does not capture the more
qualitative benefit of using the RA classifier, which is able to filter out
buildings that are hallucinated by the local segmentation in areas of high texture (as shown in Figure~\ref{fig:tree_classifier_results}).
Since buildings generally occupy only a small fraction of pixels, the overall average improvement is significantly less than
the improvement in those specific places.

\section{A DUAL LOCAL-GLOBAL CNN FOR SEMANTIC SEGMENTATION}

\begin{figure*}[t!]
\centering
\includegraphics[width=18cm]{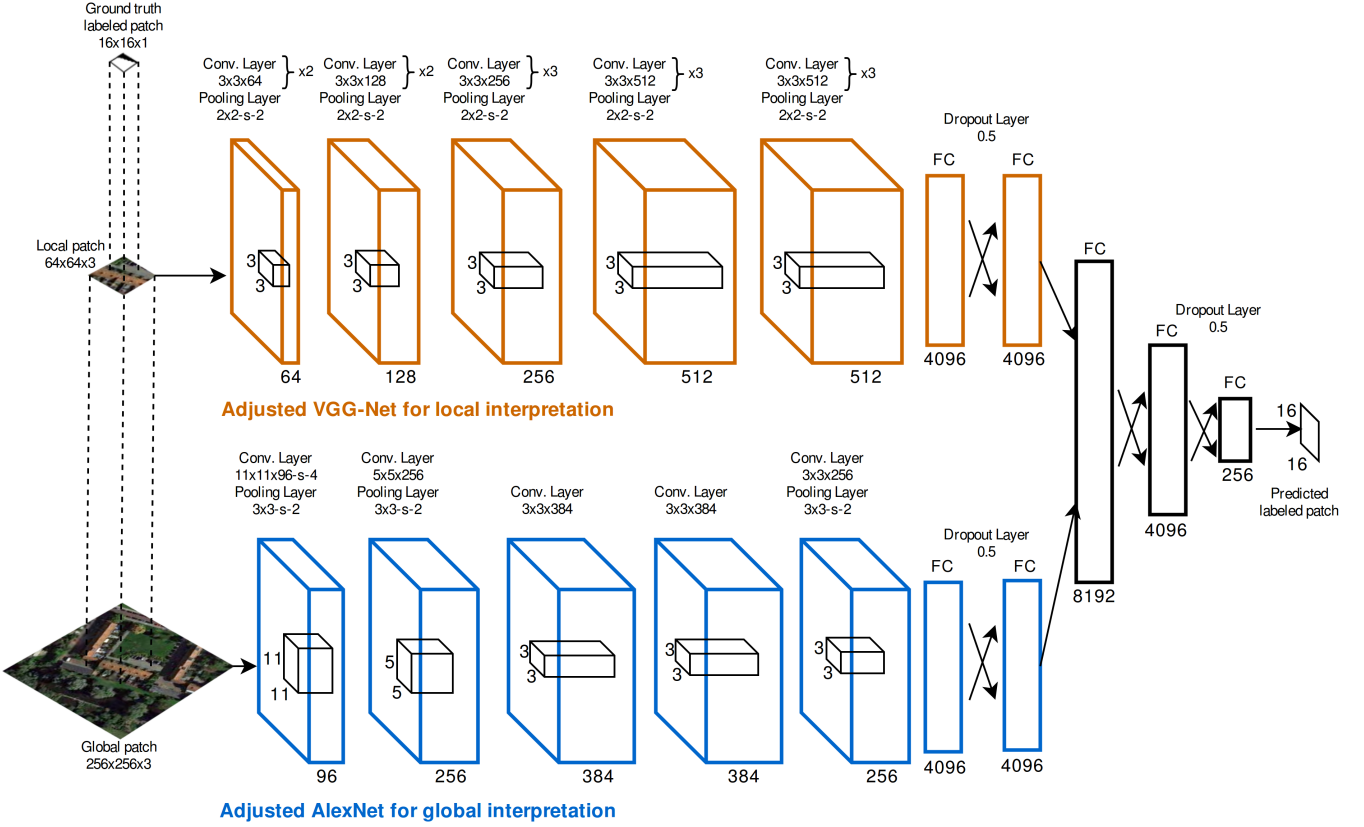}
\caption{Our proposed, dual-stream, local-global architecture LG-Net. It is formed by modifying and joining two state-of-the-art deep nets, namely VGG-Net - used here for local image interpretation ($L-Seg$) and AlexNet - used here for global interpretation of the contextual scene ($G-Seg$). Note that the L-Seg network is deeper but narrower with
smaller filter sizes (and smaller input in our case) and it is thus better suited for more detailed local processing. G-Seg network, which is shallower (fewer layers) but wider (larger input and filters), takes into consideration more information at once and it is thus more appropriate for global processing of larger areas.
The two pathways meet in the final FC layers, which combine information about object and context into a unified and balanced
higher level image interpretation.}
\label{fig:LGSeg_Net}
\end{figure*}

We take the intuition and initial tests from the previous section
a step further and create an architecture that combines the previous two models into a single local-global deep network, termed LG-Seg (see Figure \ref{fig:LGSeg_Net}). The two pathways
process information in parallel, taking as input image patches of different sizes. Then,
the superior FC layers of each individual network are concatenated and fed into three FC layers that learn how to combine local and contextual information
at the level of semantic interpretation, after features at the FC layers in each pathway have reached a relatively high level of abstraction. At this final level we expect the object and its context to start "talking" to each other and reach a final conclusion - this level is the place where bottom-up and top-down reasoning about objects meet in order to resolve conflicts and reinforce agreements. Based on
the experiments performed with the simple tree model we want to find whether the two sub-nets (Figure \ref{fig:LGSeg_Net}) indeed learn categories
at different levels, the local one focusing more on the exact shape of individual buildings and the other classifying larger residential areas with less focus on exact localization and delineation of buildings.

Residential areas are a different category, on its own. For example, small green spaces between buildings, sidewalks, parking lots or playgrounds for children
may all be part of the residential area but they are not buildings. However, their existence is indication of the presence of buildings. Residential areas \emph{exist} over large regions and at higher level of semantic abstraction: they are regions where people live and could even form communities, well beyond the idea of houses or buildings.
While a small patch of grass or concrete could be part of a residential region, the same patch of grass or concrete, when present inside a large park or on an important street, should not be seen as part of a residential place.
This aspect of complementarity between an object, such as a building, and its scene, such as the residential area,
is exactly what we want to study: what are the two pathways learning when initializing the whole network from random weights?
Our experiments, presented in Section \ref{sec:local_global_complementarity}
confirm this fact: the two networks indeed learn to process information
in complementary ways, one distinguishing more individual houses and buildings and the other focusing on larger residential areas.

We also expect the single combined network trained end-to-end to be able to produce more accurate segmentations
over the simple tree model. Note that the tree model usually does not improve the shape of the segmentation produced by the local network, but only changes
the recognition confidence, using two different thresholds, over relatively large areas. In the classifier tree case, the residential area network outputs a single label per patch, while in the LG-Seg model they are jointly trained to segment objects - technically this is a good reason why we expect a qualitative improvements in segmentation of objects shape.


\subsection{Problem formulation and learning}

We formulate the object segmentation problem in a way that is similar to the one proposed by
Mnih et. all \cite{mnih2010learning}, as a binary labeling task, where all pixels belonging to the object of interest are 1 and all the others are 0. Let
$\mathbf{I}$ be the satellite aerial image and $\mathbf{M}$ the corresponding ground truth labeled map. The goal is to predict a labeled image $\hat{\mathbf{M}}$ from an input aerial image $\mathbf{I}$, that is to learn $P(M_{ij} | \mathbf{I})$ from data, for any location $p=(i, j)$ in the image.

We train our network to predict a labeled image patch $W(\mathbf{M}, p, w_{m})$, extracted from labeled map $\mathbf{M}$,
centered at location $p$, of window width $w_{m} = 16$,
from two aerial image patches $W(\mathbf{I}, p, w_{l})$ and $W(\mathbf{I}, p, w_{g})$, centered at the same location $p$, with a smaller size window width
$w_{l}=64$ for the local patch and a larger window width $w_{g}=256$ for the global patch.
We want to learn a mapping from raw pixels to pixel labels and use
a loss function that minimizes the total cross entropy between ground truth patches and predicted label patches.
For each forward pass during learning, LG-Seg receives as input three types of patches, the 16x16 patch from the ground truth map, the local 64 x 64 image patch
and the global 256 x 256 context patch, centered at the same point and having the same spatial resolution (see Figure \ref{fig:LGSeg_Net}).

Given a set of \textit{N} examples let $\hat{\mathbf{m}}^{(n)}$ be the predicted label patch for the \textit{$n^{th}$} training case
and $\mathbf{m}^{(n)}$  the ground truth patch. Then our loss function $L$ is:

\begin{equation}
\label{eq:learning}
L=-\sum\limits_{n=1}^N \sum\limits_{p=1}^{w_{m}^{2}} (m_{p}^{(n)} \log \hat{m}_{p}^{(n)} + (1 - m_{p}^{(n)}) \log (1 - \hat{m}_{p}^{(n)}))
\label{distutv}
\end{equation}

\paragraph{Technical details:}
The minimization of this loss is solved using stochastic gradient descent with mini-batches of size 10, momentum set to 0.9, start learning rate of 0.0001 and $L_{2}$ weight decay of 0.0005. We initialize the weights using the Xavier algorithm, in order to deal with the problem of vanishing or blowing up
weights during learning in deep networks - this method automatically determines the scale of the initial weights based on the number
of input and output neurons, in order to keep the weights within a reasonable range.
All our learning and testing was ran on GPU GeForce GTX 970, with 4GB memory and 1664 CUDA cores.
Our models were implemented, trained and tested in Caffe~\cite{jia2014caffe}.

\section{EXPERIMENTAL ANALYSIS}
\label{sec:experiments}

We perform experiments on finding buildings and roads on
three datasets from different regions in the world: USA, Western Europe and
Romania. These datasets vary greatly in terms of quality and content.
\paragraph{Evaluating our models:} For the evaluation of each model, we used a qualitative measure as well as a quantitative one. The model is trained such that at a forward-pass through the network it outputs a probability for each pixel, highlighting the areas in which the classifier has a high confidence in the building prediction. The qualitative metric of evaluation involves a visual representation of the detected buildings.

In the case of quantitative evaluation of the models, the most frequently used metric for the evaluation of detection systems is the precision-recall curve. In the remote sensing literature, precision and recall are also known as correctness and completeness. It is common practice to evaluate high resolution data detectors using a relaxed version of these measures ~\cite{wiedemann1998empirical}. The relaxed version of correctness represents the fraction of predicted building pixels that are within $\rho$ pixels of a true building pixel, whilst the relaxed completeness represents the fraction of true building pixels that are within $\rho$ pixels of a predicted building pixel. The true building values are selected from the ground truth. We call $\rho$ the relaxed parameter.

\paragraph{Detection of Massachusetts Buildings:}
We start by experimenting with the relatively recent Massachusetts Buildings Dataset~\cite{MnihThesis}.
It consists of 151 high quality aerial RGB images of the Boston area. They are of size 1500 x 1500, at resolution 1 square meter per pixel, and represent
mostly urban and suburban areas, containing larger buildings, individual houses and sometimes even garages.
The entire dataset covers roughly 340 square kilometers.
It is randomly divided in a set of 137 images used for training, 4  used for the validation of the model and 10 images for testing.
We extracted approximately 700K patches from the training images and trained our model over 13 epochs for about 4 days on the GeForce GTX 970.
For computing the maximum mean F-measure over the testing set we applied the same relaxation of 3 pixels used by the competitors:
for a given classification threshold, a positively classified pixel is considered correct if it is within 3 pixels from any positive
pixel in the ground truth map. This relaxation provides a more realistic evaluation,
as borders of buildings in ground truth are often a few pixels off.

\begin{table}
\begin{center}
\caption{Results on Massachusetts Buildings Dataset.}
\label{tab:Mass_Results}
\begin{tabular}{lccc}
\hline
\rule{0pt}{10pt}
 Method&Mnih et al. ~\cite{mnih2013machine}&\multicolumn{1}{c}{Saito et at. ~\cite{saito2015building}}&\multicolumn{1}{c}{Ours}
\\
\hline
\\[-6pt]
F-measure &0.9211&0.9230& \textbf{0.9423} \\
\hline
\end{tabular}
\end{center}
\end{table}

\begin{figure}[t!]
\centering
\includegraphics[height=3.4cm]{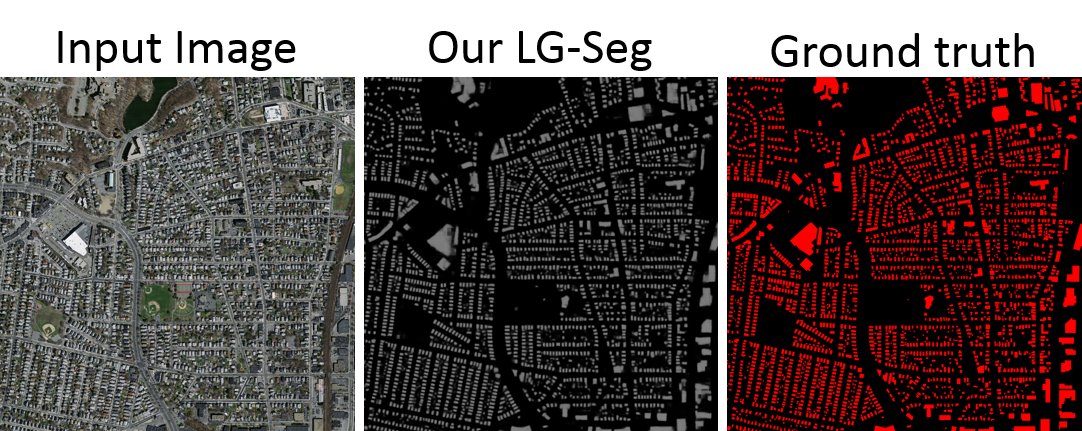}
\caption{Example of buildings detection results on the Massachusetts Dataset. Note the high level of regularity of buildings and roads,
which look very similar to each other. This permits the deep nets to learn almost perfectly and almost match human performance. Best viewed on the screen.}
\label{fig:Mass_Results}
\end{figure}

\begin{figure}[h!]
\centering
\includegraphics[height=6cm]{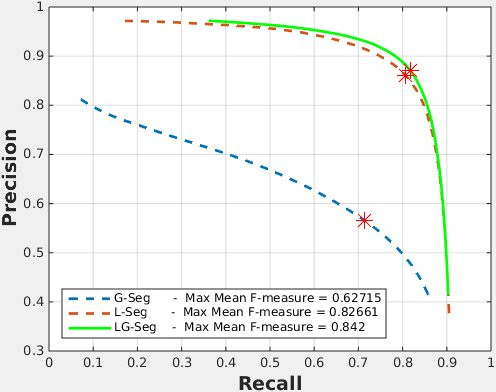}
\caption{Performance comparisons on the European Buildings Dataset between the local L-Seg model (red dotted line),
the global G-Seg model (blue dotted line) and the combined LG-Seg. Note that LG-Seg is superior, with over $1.5\%$
improvement in F-measure, on average, over L-Seg. The improvement is significant especially in regions of low
residential density where the local model tends to hallucinate buildings.
Note that G-Seg does poorly by itself as it cannot capture fine segmentation details, but it becomes valuable, as a scene processing pathway,
within the LG-Seg framework.
}
\label{fig:European_PR}
\end{figure}

\begin{figure*}[t!]
\centering
\includegraphics[width=18cm]{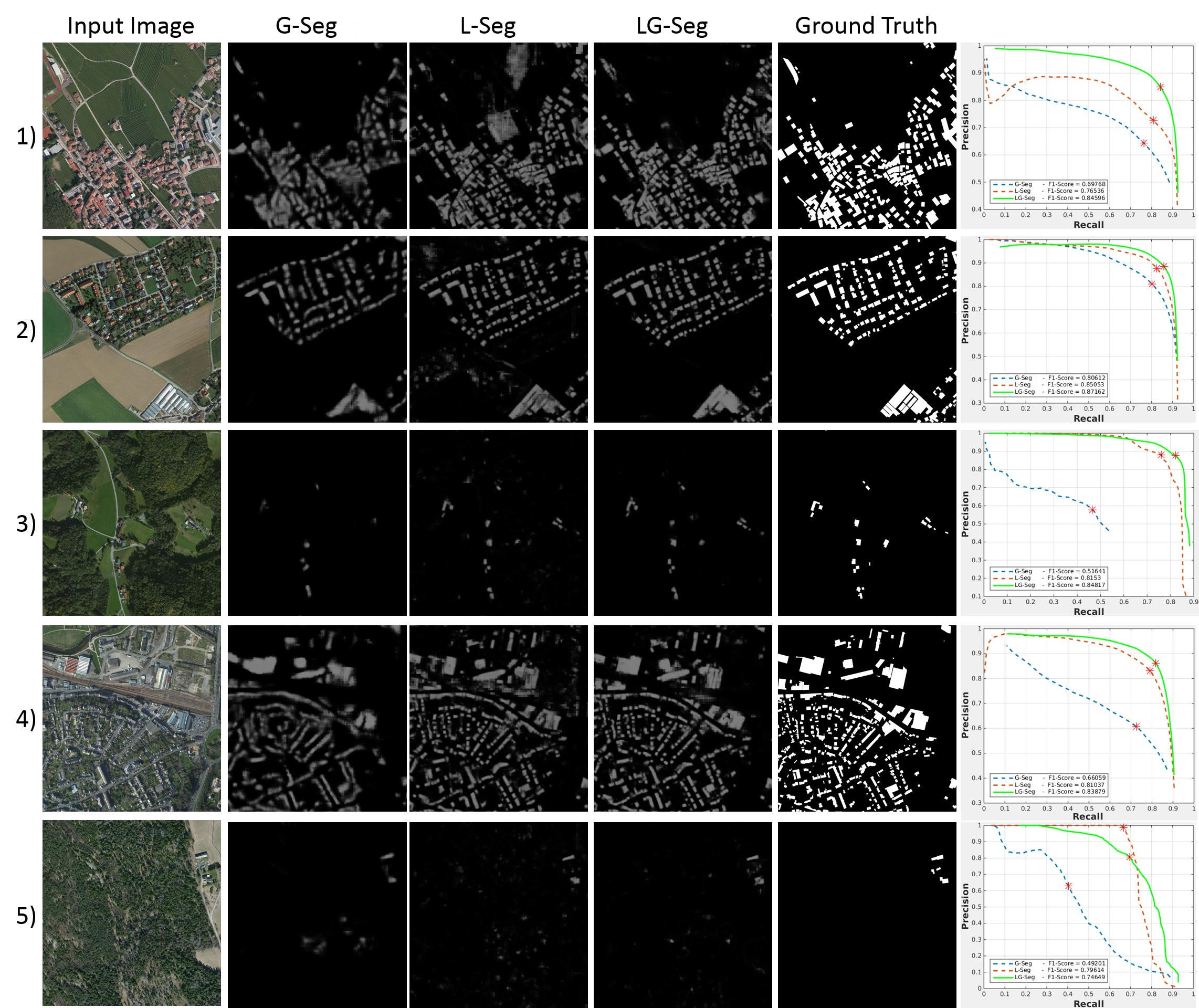}
\caption{Comparisons between the local L-Seg, global G-Seg and local-global LG-Seg architectures.
LG-Seg performs the best. By reasoning over a larger area LG-Seg is able to remove false positives (see e.g. 1). Note that LG-Seg is also able to produce more accurate building shapes (see e.g. 2).}
\label{fig:LG_comparisons_Europe}
\end{figure*}

\paragraph{Detection of European Buildings:}
Next we tested on the European Buildings Dataset, which we
collected from Western European urban and suburban areas. They contain
a lot more variation than in US,
in terms of general urban structure and roads, architecture style, layout of green spaces vs. residential areas and geography.
We have gathered 259 RGB satellite images from Google and Bing maps, of size 1550 x 1600 pixels, of resolution of about 0.8 square meters per pixel,
with locations picked randomly from different Western European countries.
Covering a larger total area of 348.5 square kilometers of urban and rural regions spread across Europe,
these images also had a lot more variation in illumination as compared to those from Boston.
We randomly selected 144 images for training (198.2 square kilometers), 10 for validation (21.3 square kilometers) and 100 for testing (129 square kilometers).
The ground truth labeled map for each individual image was generated using data from the OpenStreeMap (OSM) project. We automatically aligned the satellite images with their corresponding maps from OSM, which has manually annotated buildings. For training we extracted about 1 million patches.
We tested three models (Table \ref{tab:European_results}, Figures \ref{fig:European_PR} and \ref{fig:LG_comparisons_Europe}):
our full LG-Seg net, and models formed by keeping only one pathway, G-Seg with  the adjusted AlexNet only and
L-Seg formed by the adjusted VGG-Net only. We wanted to test the capabilities of each separately and study the potential
advantage of combining them into a single LG-Seg. All models were trained until complete convergence of the loss, with the G-Seg model taking $34$
epochs, L-Seg model $23$ epochs and LG-Seg converging the fastest, in only $12$ epochs. Training time varied between 3 to 6 days on our
GeForce GTX 970.

\begin{table}
\begin{center}
{\caption{Results of our trained models on the European Buildings Dataset.}\label{tab:European_results}}
\begin{tabular}{lccc}
\hline
\rule{0pt}{10pt}
 Method&{G-Seg}&\multicolumn{1}{c}{L-Seg}&\multicolumn{1}{c}{LG-Seg}
\\
\hline
\\[-6pt]
F-measure &0.6271&0.8266& \textbf{0.8420}\\
\hline
\end{tabular}
\end{center}
\end{table}

\begin{figure}[h!]
\centering
\includegraphics[height=11.5cm]{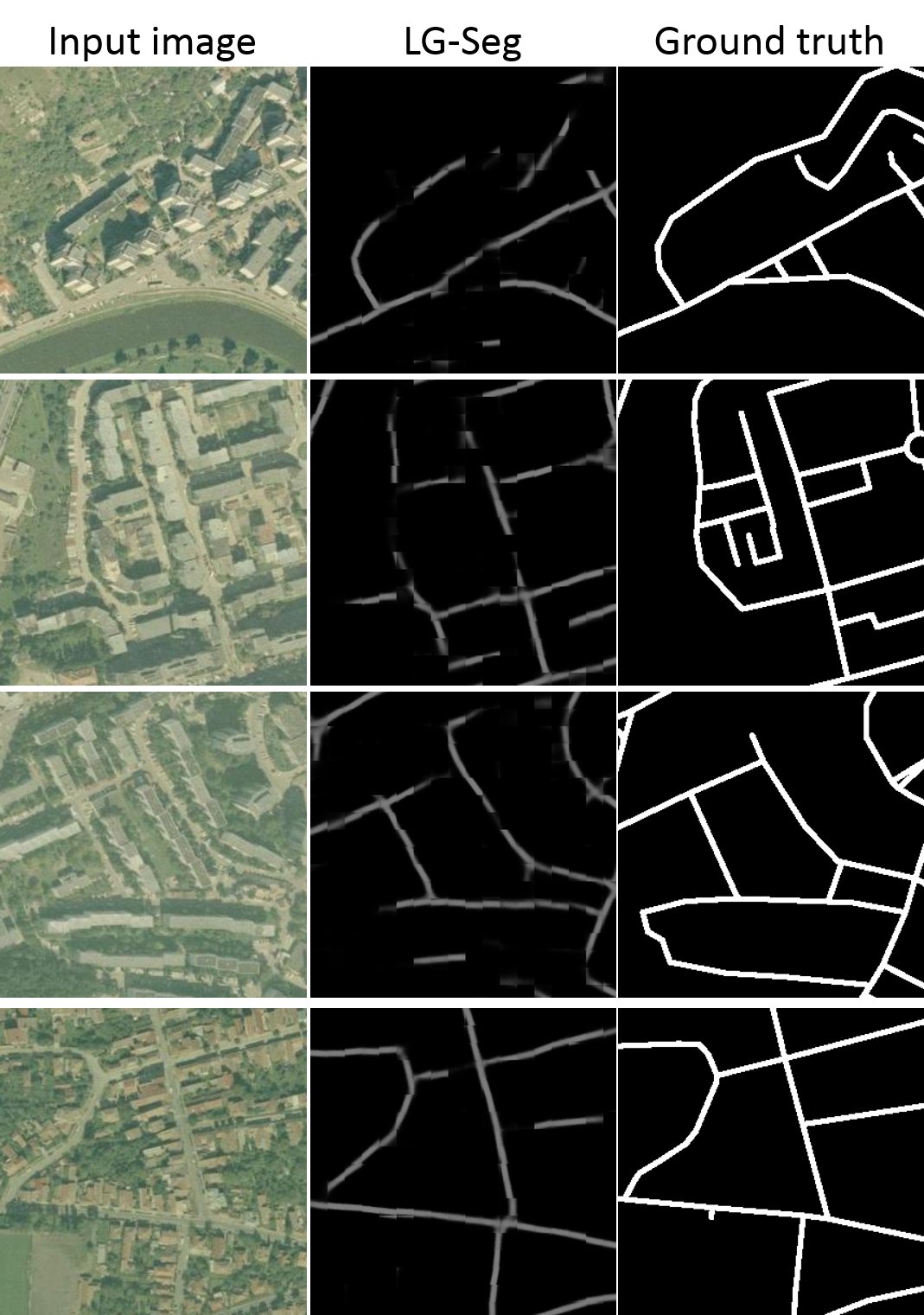}
\caption{Example results on Romanian roads. Note how difficult the  task is on these images, posing
a real challenge even for humans.}
\label{fig:Romanian_roads}
\end{figure}

\paragraph{Detection of Romanian Roads:}
We have collected aerial images of two Romanian cities, Cluj and Timisoara, of size 600 x 600 and resolution 1 square meters per pixel and automatically aligned them
with OSM road maps to obtain the ground truth labels.
For Cluj we have 3177 images covering an area of about 70 square kilometers, and for Timisoara 4027 images for an area of 72 square kilometers. Images have significant spatial overlap,
such that there is one image for each road intersection (as estimated from OSM).
For this dataset we trained our model on the task of road detection, as roads are the only category represented relatively well in OSM
over these Romanian regions. We used Timisoara images for training our LG-Seg
model and Cluj images for testing.
This dataset offers a different task, that of road detection, and also
a much more challenging one due to limitations and variations in the data.
Different from the other image sets,
this one is of significantly lower quality, with large variations in the road structure, their type, width and length. Moreover, often the roads are completely occluded by trees and the OSM road maps do not match correctly what is seen in the image (see examples in Fig. \ref{fig:Romanian_roads}).
Also note that Timisoara and Cluj have different urban styles, which
brings an extra degree of difficulty for learning and generalization.
For these many reasons, on this dataset, the
problem of recognition is tremendously difficult and pushes the limits of deep learning to a next level,
as reflected by the significantly lower performance.

In Table \ref{fig:Romanian_roads} we present results and comparisons between the LG-Seg and L-Seg models
on the Romanian Roads Dataset. Again, both quantitatively and qualitatively the LG-Seg model wins.
In this particular case, the L-Seg model had the advantage of being
fully pre-trained on a much larger set of images, covering about 775 square kilometers from Romania, of higher quality and resolution (collected from Google and Bing Maps) and then fine-tuned on our Timisoara set. The LG-Seg model was only trained on Timisoara images. Qualitative results on this set are shown in Figure \ref{fig:Romanian_roads}.
The examples show the high level of difficulty posed by this challenging dataset, which we
make available for download~\footnote{https://sites.google.com/site/aerialimageunderstanding/}.
We believe it poses a very challenging task and could help in new valuable research in
aerial image understanding.

Our experiments on the three datasets, of different content and quality, reveal one more time the
importance of data in learning. When the structures are regular and look very similar across images, such as it is the case
with the Massachusetts Buildings, the performance reaches almost human level. However, as the variations in the data, lack of image
quality and frequency of occlusions increase, the performance starts degrading, dropping by almost $30\%$ on the Romanian roads.
These results prove that
aerial image understanding is far from being solved even in the context of state-of-the-art
deep networks and that it remains a very challenging problem.

\begin{table}
\begin{center}
{\caption{Results on Romanian Roads Dataset}\label{tab:Romanian_roads}}
\begin{tabular}{lcc}
\hline
\rule{0pt}{10pt}
 Method&{L-Seg}&\multicolumn{1}{c}{LG-Seg}
\\
\hline
\\[-6pt]
\quad F-measure & 66.1\% & \textbf{66.5}\% \\
\hline
\end{tabular}
\end{center}
\end{table}

\begin{figure*}[t!]
\centering
\includegraphics[width=18cm]{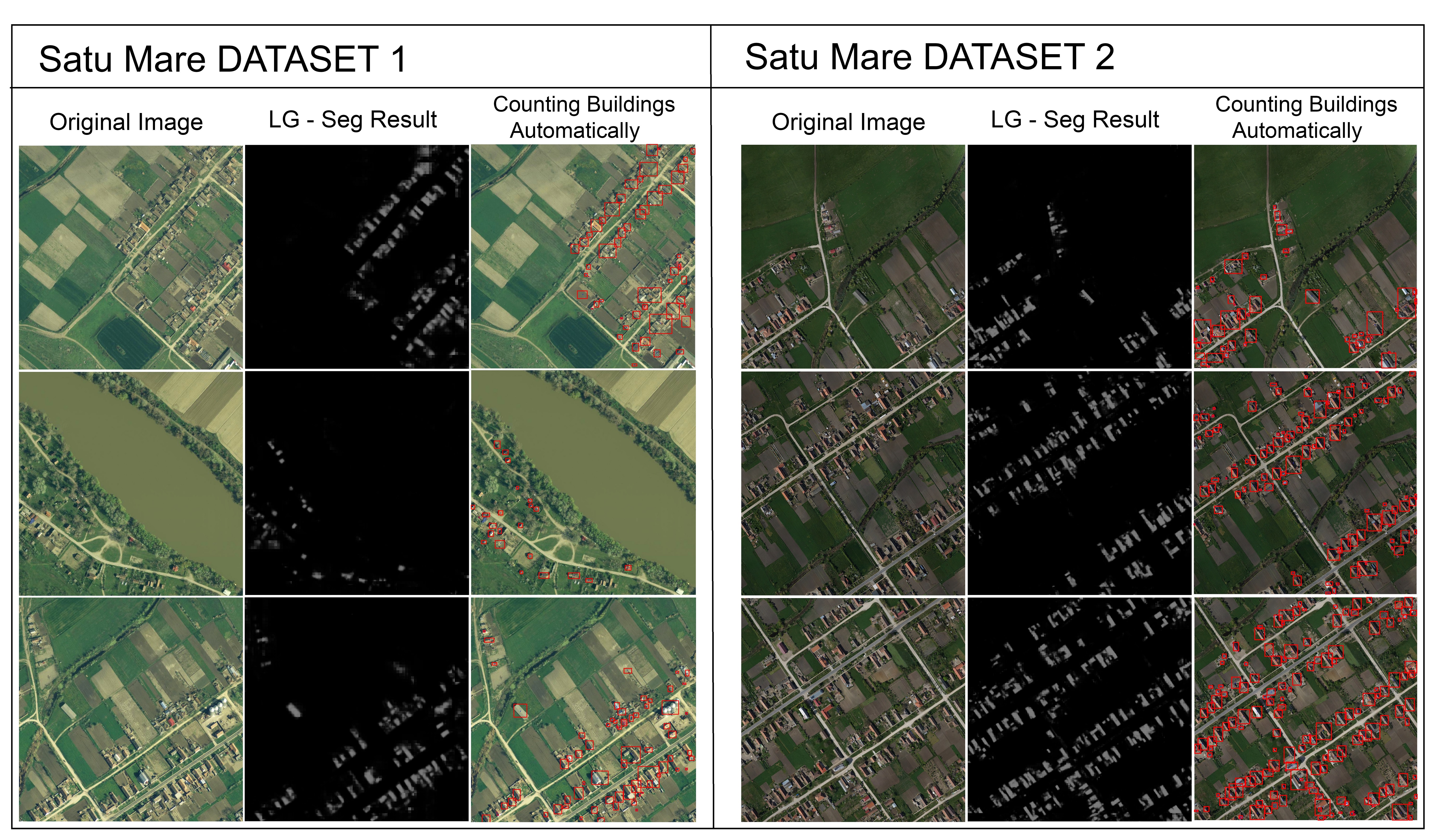}
\caption{\textbf{Qualitative results on detecting and counting houses}.
Results are shown on two different datasets. The differences between our datasets can be easily spotted. In the first dataset images have low buildings
density at low resolution, while the second dataset contains higher quality images and higher density of houses.
For each dataset we show three example results. On the first column we show the input RGB images, on the second column the prediction map of our LG-Seg model
and on the third column original images with the detected house bounding boxes in red overlaid.
The quantitative results of house detection and counting are
presented in Table ~\ref{tab:SM_Statistics}.
}
\label{fig:SM_Counting_Buildings}
\end{figure*}

\begin{table*}[t!]
\begin{center}
{\caption{House Detection and Counting Statistics on Satu Mare Datasets}\label{tab:SM_Statistics}}
\begin{tabular}{lcccccc}
\hline
\rule{0pt}{10pt}
 \textit{Satu Mare Dataset 1}&{Human count}&{Machine count}&{TP}&{FP}&{FN}&{Residential}
\\
\hline
\\[-6pt]
\quad   & \textbf{156} & \textbf{137}& 106 & 18 & 33 & 13  \\
\hline
\textit{Satu Mare Dataset 2}&{Human count}&{Machine count}&{TP}&{FP}&{FN}&{Residential}
\\
\hline
\\[-6pt]
\quad   & \textbf{295} & \textbf{290}& 239 & 31 & 18 & 20  \\
\hline
\end{tabular}
\end{center}
\end{table*}

\subsection{Detection and Counting of Houses}
An interesting task that is also useful in applications such as real estate and cadastre mapping, urban planning and landscape monitoring, is the detection
and counting of houses within a given area. For this experiment we have collected images from different areas around the city of Satu Mare, Romania,
thus creating two new datasets, different from those presented in the previous sections. Besides the fact that the images were collected from rural regions
with lower house densities, they were also retrieved at different spatial resolutions. For these images we have not used pixel-wise ground truth labels
for training and evaluation,
since these regions were not properly labeled on OSM. We refer to these datasets as Satu Mare 1 and Satu Mare 2, and
make them available for download at~\footnote{https://sites.google.com/site/aerialimageunderstanding/}.
The datasets as well as the experiments are presented next:

\paragraph{Satu Mare Dataset 1:} This dataset represents an aerial map of size 20000 x 20000 and spatial resolution of 0.5 x 0.5 square meters per pixel. It was
divided in 400 tiles of size 1000 x 1000 pixels.  The tiles were then resized with a re-scale factor of 1/2 in order to bring the images at a resolution of 1m per pixel,
closer to the one that our LG-Seg model was trained on (the European Buildings dataset). Note that the Satu Mare images were only used at test time,
without any fine-tuning of the LG-Seg model. We expect that such refinement would have increased performance. However, even for this case, our results, presented next,
are very promising. Also note that the houses from this region are sparsely placed,
with relatively few residential areas and large vegetation regions. Also, the images
are of poorer quality (see Figure~\ref{fig:SM_Counting_Buildings}) than those from  the European dataset that was used for training. This
makes building detection a difficult task even for humans.

\paragraph{Satu Mare Dataset 2:} A different aerial map of size 20000 x 20000 and spatial resolution of 0.05 x 0.05 square meters per pixel was divided in 4 tiles of size 10000 x 10000. In this case we applied a re-scale factor of 1/20 in order to bring the images to 1m per pixel.
Also different from Satu Mare 1 dataset, the buildings in Satu Mare 2 are more tightly clustered together, with a larger variation in house density.

\paragraph{Estimating the number of houses:} We applied a post-processing method to the output of our model. We obtained hard 0-1
prediction maps by applying the threshold corresponding to the optimal F-measure learned from the European Buildings dataset.
Next, we apply a morphological erosion operation in order to separate closely placed buildings. Each connected component
obtained in this manner is considered to be a separate house. These connected components of white pixels
are then used for estimating the number of houses and also
for estimating the bounding box and shape of each individual house.
Qualitative results of our method can be viewed in Figure ~\ref{fig:SM_Counting_Buildings}.

In order to obtain an approximate estimate of the accuracy of our method on house detection and counting, we randomly selected images
covering an area of approximately 3 sq Km, from each dataset. Then we manually selected bounding boxes for each house present in the image, in order to compare
against the automatically detected ones.
We present the results in Figure~\ref{fig:SM_Counting_Buildings} and the quantitative evaluation in
Table~\ref{tab:SM_Statistics}. The results show that we are able to give house counts that are very close to the human estimates.
In some cases a clear separation of the individual houses was not possible, as they were very close to each other, forming a small residential region.
We mark these boxes, which contain at least two houses, as Residential in Table~\ref{tab:SM_Statistics}.

The experiments are encouraging: on the Satu Mare 1 images,
out of 156 human-labelled houses, our system was able to detect correctly 106 of individual houses (with an overlap over union of bounding boxes
greater than 0.5), while 13 bounding boxes correctly contained residential regions (groups of houses).
Knowing that for each residential hit there are at least two houses, we can compute an approximate value of the precision of our system (88\%), and recall (80\%). On the second dataset, the one with high-density housing and better spatial resolution and quality of images,
out of a total of 295 manually-labelled houses, there were 239 correctly detected ones, resulting in a precision of 90\% and recall of 94\%. As expected, higher image quality improves the building detection rate using our trained model. Note, however, that even in low resolution conditions the system offers promising results.

\subsection{Discussion on Local-Global Complementarity}
\label{sec:local_global_complementarity}

\begin{figure*}[t!]
\centering
\includegraphics[width=15cm]{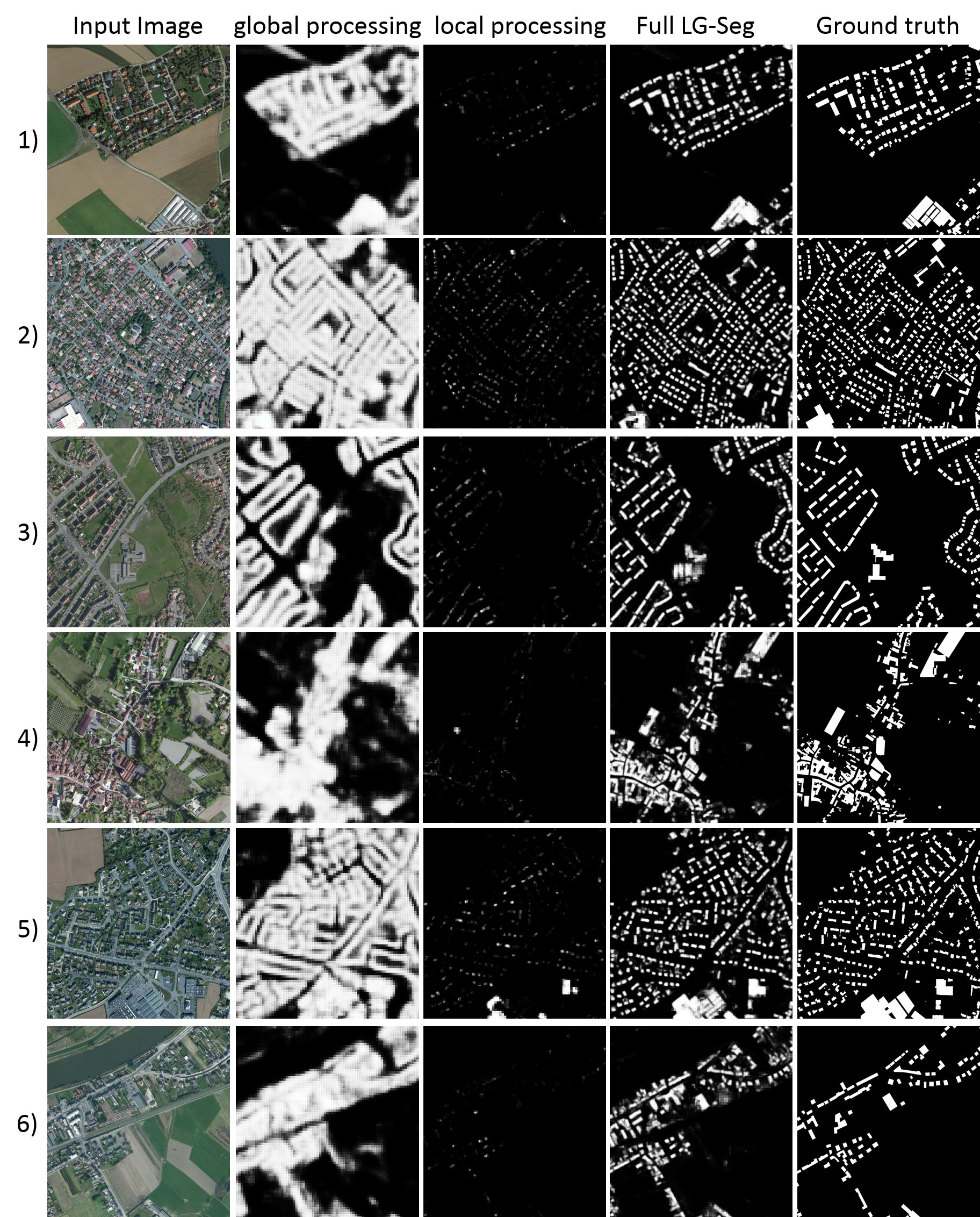}
\caption{In these experiments we aim to find what the two pathways have learned. The second column shows results when only the global pathway is fed with real image signal, the other being given blank image as input. The third column shows the opposite case, when only the local pathway is given real information. The fourth column presents the output of the network running normally, with both pathways having image input. Note that the global sub-net learns to detect residential areas similar to our initial classifier for such regions. Example 6) shows the results of our model on the same image as in Figure 1.B. Note that the residential area segmentation produced by the LG-Seg is superior to the one produced by the RA classifier, even though in the case of LG-Seg it was not asked to learn about residential areas.
Also note that the local pathways focuses only on small, detailed structures. The imbalance between the energy levels of the outputs is due to the fact that one of the inputs is blank, thus unbalancing the way energy flows at the highest FC layers. The results also suggest that the two pathways have roles of both reinforcement and inhibition. For example the local pathway will inhibit the global positive outputs in spaces
between buildings, whereas the global pathway will inhibit the local hallucinations in
areas of low residential density. We can safely conclude that the two pathways work in complementarity.}
\label{fig:Local_Global_Complements}
\end{figure*}

One question that arises in our experiments with the dual-stream architecture is what are the two pathways learning?
What is their individual role in the combined output? Our intuition
was that they probably learn complementary ways of processing data.
We intentionally chose two different types of networks with different image region sizes as input,
in order to encourage different learning
along the two pathways. We hoped for two sub-nets with complementary ways of "seeing" the scene -  similar to the initial CNNs, one for
residential area classification (RA) and the other for local detection and segmentation of buildings (L-Seg).

We designed a set of experiments in order to better understand the role of each sub-net. After training the full LG-Net, we performed the following: first, we ran the model over the test images by providing the local pathway with the correct image input, but giving a blank image to the global pathway. The blank image was the average of the original input image, for each RGB channel averaged separately.
Then, we performed the opposite experiment and switched the inputs, by giving the original image to the global subnet and blank images to the local one. The idea was to see how, in the fully trained model, each path contributes to the final decision.

The results, presented in Figure~\ref{fig:Local_Global_Complements} are both very interesting and satisfying. When provided with information for local processing only, the network responds only to small buildings with very clear structure, having crisp, very local responses over individual houses or buildings.
On the other hand, when given information only to the global sub-net, the network produced a result that was closer to a soft residential area segmentation, in which individual buildings were undistinguishable from each other - a result, very close, but of higher quality than our initial residential area detection based on the same adjusted AlexNet architecture.

What makes these results really interesting is the fact that we did not tell these two pathways to take these different roles - all we did is choose two different architectures, gave them two different image sizes as input and let them learn, from random initializations, by themselves within the joint network. Complementarity, which was our main goal when starting this work, was learned automatically by our model from scratch.

\section{CONCLUSIONS}

We have studied different ways of combining local appearance and global contextual information
for semantic segmentation in aerial images. After testing initial simpler models that proved the usefulness
of reasoning about visual context in object detection and segmentation, we have proposed a novel dual local-global network
which learns completely by itself to look at objects from two complementary perspectives. When given the task of segmentation of buildings
the network learns to treat each pixel, in parallel, both as a part of a building and as a part of a larger residential area. It also learns to
combine the two reciprocal views in a harmonious way during the final layers of processing, before providing the final result.
Our experiments on the
roads dataset also emphasize how difficult aerial image understanding still is,
even for high performance, state-of-the-art deep neural networks,
especially
in cases of poor lighting, low image quality, occlusion and
high degree of variations in objects structure and shape.
We believe that these limitations will be overcome by the usage of context at even higher levels of abstraction and
reasoning. Consequently, we see our work as having the potential to influence future research that will
shed new light on the
understanding of context in vision.

\ack The authors would like to thank Dragos Costea for his dedicated assistance with
some of our experiments.


\bibliography{ecai}
\end{document}